\documentclass{article}

\usepackage[preprint]{neurips_2026}

\usepackage[utf8]{inputenc} 
\usepackage[T1]{fontenc}    
\usepackage{hyperref}       
\usepackage{url}            
\usepackage{booktabs}       
\usepackage{amsfonts}       
\usepackage{nicefrac}       
\usepackage{microtype}      
\usepackage{xcolor}         

\usepackage{amssymb,amsmath,amsthm,latexsym}
\usepackage{graphicx}
\usepackage{sg-macros}
\usepackage{enumitem}

\usepackage{algpseudocode}
\usepackage{algorithm}
\algrenewcommand\alglinenumber[1]{\textcolor{gray}{\tiny #1:}}

\usepackage{pgfplots}
\pgfplotsset{compat=1.5, every axis/.append style={font=\small, /pgf/number format/1000 sep={}}, every mark/.append style={solid},}
\usepackage{tikz}
\usetikzlibrary{arrows.meta, angles, math, calc, shapes, intersections, positioning}

\title{Back from the Future: Key-Value Cache Management by Counter-Causal Surprise}

\author{%
	Stephen Gould\textsuperscript{1,2} \qquad
	Anton van den Hengel\textsuperscript{1,3} \\[0.4em]
	\normalsize
	\textsuperscript{1}Metacognition AI \quad
	\textsuperscript{2}Australian National University \quad
	\textsuperscript{3}Adelaide University \\
	\texttt{\{steve,anton\}@metacognitionai.com}
}

\begin{document}
\maketitle

\begin{abstract}
	Key-value (KV) cache management through compression and eviction strategies has emerged as an important research direction in recent years. Computational demands of large language models (LLMs) and their multi-modal variants during output generation can be partially alleviated by caching previous key and value calculations needed by subsequent scaled dot-product attention operations. However, this leads to another problem: the size of the resulting KV cache grows linearly with context length and quickly consumes all available GPU memory when either the prompt or the generated output are long. KV cache management periodically prunes entries from the cache thereby reducing its memory footprint while attempting to retain sufficient information for accurate generation. A by-product is faster inference speed. We propose a simple yet effective KV eviction scheme motivated by the insight that past tokens which can be well-predicted from more recent tokens are redundant and their associated keys and values can be removed from the cache. To score entries for eviction we run the model on the tokens in their original order, reusing the key and value representations already stored in the KV cache, and applying a counter-causal attention mask so that each position attends only to its future context. This is in-distribution, tied directly to the actual cache contents, and requires no additional training. To further reduce cost, we additionally propose a fast single-layer approximation that restricts the counter-causal pass to the last transformer layer, achieving a significant speedup per refresh cycle at marginal accuracy cost. We evaluate our strategy on various open-source LLMs and benchmark datasets showing competitive or improved performance over other state-of-the-art methods.
	Reference code is available at \url{https://github.com/metacognitionai/counter_causal}.
\end{abstract}

\section{Introduction}
\label{sec:intro}

A pivotal innovation in the deployment and adoption of large language models (LLMs) was the introduction of the key-value (KV) cache. By providing a mechanism for storing previously computed keys and values needed for scaled dot-product attention within the transformer architecture~\citep{Vaswani:NIPS2017}, the KV cache significantly reduces compute thereby enabling LLMs to scale to long context windows during inference. However, these computational savings come at a cost. The KV cache consumes a sizable memory footprint and as the context windows grow beyond a few thousand tokens the GPU memory required to store the cache becomes prohibitive. Consequently, there has been considerable research effort in finding strategies for reducing the memory footprint of the KV cache in recent years.

One line of research focuses on evicting entries from the KV cache while attempting to maintain the quality of the generated output. An obvious approach is to remove the oldest entries in a sliding window fashion once the KV cache reaches some pre-defined maximum size~\citep{Beltagy2020Longformer}. Another approach is to compress the input prompt before providing it to the model thereby reducing the number of tokens in the context window before the model starts generating output~\citep{Li:EMNLP2023, Fei:ACL2024}. Clearly this only works when the subsequent text generated by the model is small, e.g., answering multiple-choice questions or prompts that require short answers, or when applied episodically such as in a chain-of-thought (CoT) reasoning pipeline~\citep{Xia:EMNLP2025}.

More sophisticated approaches dynamically try to identify the most influential entries in the cache to keep and remove irrelevant ones~\citep{Zhang:NIPS2023, Oren:arXiv2024, Cetin:ICLR2025}. These typically work by analyzing the attention matrix computed during output generation and argue that keys receiving high attention scores are influential and should be kept (along with their corresponding values). The tokens associated with these keys have become known as heavy hitters in the literature~\citep{Zhang:NIPS2023}. However, these attention-based scores are susceptible to a \emph{self-reinforcing bias}: tokens retained in the cache continue to receive attention from subsequent decoding steps, increasing their cumulative score and making them even more likely to be retained in future eviction cycles. Conversely, a factually critical but low-frequency token mentioned once may attract too little attention to survive eviction, even though it may be essential for answering a later query. We demonstrate this failure mode empirically in Section~\ref{sec:expr} for where accuracy of attention-based methods degrade more rapidly on some task/model/cache-size combinations.

In this paper we propose a novel eviction strategy based on the observation that if we are able to predict a past token from the tokens that follow it, then the information provided by that past token has already been embedded somehow in the later context. To capture this idea, we define a quantity called \emph{counter-causal surprise} that measures the merit in keeping key-value pairs on how well or poorly their associated tokens can be predicted. Tokens with low surprise provide little further usefulness to the generation process and can be removed from the KV cache. Tokens with high surprise, on the other hand, cannot be predicted by the tokens that currently follow them. As such they may contain important unique information for output yet to be generated by the LLM, and so should be kept. This strategy has the added benefit of preferring more recent tokens, which helps the LLM focus on the latest context similar to the sliding window approach but in a more principled manner. Importantly, our method does not require any training and can be used by any model performing auto-regressive token generation that uses a KV cache.

Our mechanism for scoring whether to evict KV cache entries runs a forward pass in the original token order, reusing the cached keys and values already computed during generation, and applying a counter-causal (upper-triangular) attention mask so each position attends only to future tokens in the cache. This costs $O(L n^2)$ per refresh where $L$ is the number of transformer layers and $n$ is the current cache size, but can be done on a batch of tokens simultaneously. We additionally propose a fast single-layer approximation that restricts the counter-causal pass to the last transformer layer only, reducing per-refresh cost to $O(n^2)$ and achieving 7--9 times speedup (as low as 7.9ms per refresh at a 512-token cache size on Qwen2.5-7B).

We conduct experiments on several benchmark datasets and two families of open-weight large language models of different sizes. The benchmarks cover tasks with either long detailed prompts or requiring long-form generated text. This allows us to test our method during both the pre-fill phase and decode (generation) phase of LLM inference. Results show that our method is competitive against or outperforms existing methods.

\section{Background and Related Work}
\label{sec:background}

The transformer architecture underpins modern large language models (LLMs) and their multi-modal extensions, enabling arbitrary length sequence modeling through scaled dot-product attention~\citep{Vaswani:NIPS2017}. Contemporary models such as Qwen2.5 and LLaMA 3.1, used in our experiments, perform auto-regressive decoding, where previously computed key-value (KV) pairs are cached to avoid redundant computation~\citep{qwen2025qwen25technicalreport, grattafiori2024llama3herdmodels}. While essential for efficient inference, the KV cache grows linearly with sequence length (and attention layers), creating both memory and bandwidth bottlenecks, particularly for long-context generation.

Beyond growth concerns, recent work at the system level has focused on improving how KV caches are stored and accessed. For example, paged or virtualized KV cache designs (as popularized in vLLM-style systems~\citep{Kwon:SOSP2023}) treat the cache as a form of virtual memory, enabling efficient batching, prefix sharing, and reuse across requests. At the algorithmic level techniques such as FlashAttention~\citep{Dao:NIPS2022} efficiently moves data across memory units on the GPU to reduce memory reads and writes while computing attention. These approaches improve throughput and memory utilization without altering the semantic content of the cache or exactness of the attention calculation. They are complementary to methods that modify or prune stored representations, and hence to the work presented in this paper.

{\bf Persistent and Augmented Memory.}
Several works extend the KV cache with persistent or external memory to incorporate information from outside the current context window. \citet{Sukhbaatar:ICLR2020} introduce learned memory slots that persist across sequences, effectively augmenting the model's working memory. More recently, \citet{Eyuboglu:2025} propose cartridges: pre-trained KV caches derived from specialized corpora that can be loaded at inference time, reducing the need for long in-context demonstrations. \citet{Xing:ICLR2026} analyze KV cache dynamics during the pre-fill phase and adaptively switch between ``fast'' and ``slow'' reasoning modes, selectively increasing computation only for difficult inputs. Complementary work by \citet{Behrouz:2024} explores long-term neural storage using surprise-based signals to prioritize important information, although this operates in a forward-looking manner and is not strictly framed as KV cache management.

{\bf Input Context Compression.}
Another class of methods reduces the effective input length prior to or during processing. \citet{Li:EMNLP2023} identify redundancy in input sequences using self-information measures to filter lexical units, while \citet{Fei:ACL2024} segment inputs into topic-coherent chunks and summarize each segment using an auxiliary pre-trained model. These approaches operate primarily as preprocessing steps and therefore do not directly address KV cache growth during auto-regressive decoding. Other directions explore online compression, such as incrementally summarizing earlier context during generation, trading exact token-level fidelity for compact semantic representations~\citep{Ge:ICLR2024, li2024snapkv}.

{\bf Dynamic Context Eviction.}
In contrast to compression, context eviction methods explicitly manage the KV cache during generation, making them well suited to long-form decoding. \citet{Zhang:NIPS2023} propose the $H_2$O framework, which identifies heavy hitter tokens that disproportionately influence attention scores. They suggest maintaining a balance between heavy hitter and recent tokens in the cache.
\citet{Oren:arXiv2024} propose TOVA, which uses the attention weights of the most recently processed token at the last transformer layer to score cached tokens for eviction. Unlike $H_2$O, the TOVA method decouples the eviction score from accumulated generation attention, but like $H_2$O it relies on forward-pass attention and is susceptible to the self-reinforcing bias, which can be a problem when applied during the pre-fill phase where intermediate context tokens provide a poor eviction signal for multi-hop tasks.

\citet{Cai:arXiv2024} propose PyramidKV, which allocates KV cache budgets non-uniformly across transformer layers (fewer slots in earlier layers, more in later layers) based on the observation that attention patterns concentrate increasingly in later layers. This is orthogonal to our scoring criterion and could in principle be combined with counter-causal eviction in future work.
\citet{KVzip:arXiv2025} propose KVzip, which compresses the KV cache by selecting entries whose removal minimally degrades a reconstruction of the context; the method is query-agnostic and requires a secondary reconstruction pass. Like our approach it performs an extra inference step but uses a causal reconstruction objective.
\citet{Feng:arXiv2024} propose Ada-KV, which dynamically allocates per-head and per-layer cache budgets based on attention entropy, adapting the budget to the content of each input rather than using a fixed global budget. This layer-adaptive allocation is complementary to eviction scoring.
\citet{Cetin:ICLR2025} introduce neural attention memory models (NAMMs), which learn token eviction policies using features derived from attention matrices, including backward attention that allows tokens to attend to future positions. Their approach trains a lightweight model to score tokens for eviction, whereas our method instead estimates predictive utility directly and does not require additional training.

Other related approaches approximate or sparsify attention~\citep{Child:arXiv2019, Jiang:NIPS2024} rather than explicitly evicting tokens, for example through low-rank approximations~\citep{Song:CVPR2024} or selective pruning prior to full attention computation. These methods blur the boundary between architectural modification and runtime cache management.

{\bf Low-Bit Representations and Quantization.}
Orthogonal to token-level strategies, several works reduce KV cache memory through floating-point representation compression. \citet{Hooper:arXiv2024} and \citet{Zandieh:ICLR2026} explore low-bit quantization of KV tensors, significantly reducing memory footprint while attempting to preserve model quality. These methods are complementary to eviction and compression, as they reduce the memory cost per token rather than the number of stored tokens.

In short, existing approaches to KV cache management fall into four broad categories: (i) augmenting memory with persistent or external mechanisms, (ii) compressing or summarizing input context, (iii) dynamically managing the cache through eviction or sparsification, and (iv) reducing representation cost via quantitation. Algorithmic level optimizations such as FlashAttention further improve efficiency without altering model behavior. Our work builds most directly on context eviction, focusing on a principled, training-free method for estimating and preserving the most predictive elements of the cache during pre-fill and auto-regressive generation.

\section{KV Cache Management by Counter-Causal Surprise}
\label{sec:main}

\begin{figure}
	\centering
	\scalebox{0.8}{\begin{tikzpicture}[every node/.style={minimum height=10mm, inner sep=0, fill opacity=0.2, text opacity=1}]
		\node[draw, dashed, minimum width=5mm, fill=blue] (kt) at (0, 0) {$k_t$};
		\node[draw, minimum width=10mm, fill=blue, right=0cm of kt] (kprev) {$K$};
		\node[draw, minimum width=15mm, fill=blue, right=0cm of kprev] (kdash) {$K'$};

		\node[draw, dashed, minimum width=5mm, fill=red] (vt) at (0, -1.2cm) {$v_t$};
		\node[draw, minimum width=10mm, fill=red, right=0cm of vt] (vprev) {$V$};
		\node[draw, minimum width=15mm, fill=red, right=0cm of vprev] (vdash) {$V'$};

		\node[draw, dashed, minimum width=5mm, fill=yellow] (xtb) at (0, -2.4cm) {};
		\node[draw, minimum width=10mm, fill=yellow, right=0cm of xtb] (xprev) {$X$};
		\node[draw, minimum width=15mm, fill=yellow, right=0cm of xprev] (xdash) {$X'$};

		\node[draw, dashed, minimum width=5mm, fill=white] (qt) at (0, 1.2cm) {$q_t$};
		\node[draw, minimum width=5mm, fill=white, fill opacity=0.2, text opacity=1] (xt) at (-1.5cm, -0.6cm) {$x_t$};
		
		\draw (xt.east) to (xt.east -| -0.75cm, 0);
		\draw (-0.75cm, -2.4cm) edge[-{LaTeX[]}] (xtb.west) -- (-0.75cm, 1.2cm) edge[-{LaTeX[]}] (qt.west);
		\draw[-{LaTeX[]}] (kt.west -| -0.75cm, 0) -- (kt.west);
		\draw[-{LaTeX[]}] (vt.west -| -0.75cm, 0) -- (vt.west);

		\node[
			draw, rounded corners=5pt, minimum width=5cm, minimum height=1cm, fill=gray, rotate=90
		] (sdpa) at (4cm, -0.6cm) {Scaled Dot-Product Attention};

		\node[draw, minimum width=5mm, fill=white] (yt) at (5.5cm, -0.6cm) {$y_t$};

		\draw[-{LaTeX[]}] (qt.east) to (qt.east -| sdpa.north);
		\draw[-{LaTeX[]}] (kdash.east) to (kdash.east -| sdpa.north);
		\draw[-{LaTeX[]}] (vdash.east) to (vdash.east -| sdpa.north);
		\draw[-{LaTeX[]}] (sdpa) to (yt);

		\node[draw, dashed, minimum width=10mm, fill=white, right=7cm of kprev] (rkprev) {$K$};
		\node[draw, minimum width=15mm, fill=white, right=0cm of rkprev] (rkdash) {$K'$};

		\node[minimum width=3mm, fill=blue, left=-4mm of rkprev] {};
		\node[minimum width=1mm, fill=blue, left=-9mm of rkprev] {};
		\node[minimum width=1mm, fill=blue, left=-7mm of rkprev] {};
		\node[minimum width=2mm, fill=blue, right=2mm of rkprev] {};
		\node[minimum width=1mm, fill=blue, right=5mm of rkprev] {};
		\node[minimum width=2mm, fill=blue, right=8mm of rkprev] {};
		\node[minimum width=4mm, fill=blue, right=11mm of rkprev] {};

		\node[draw, dashed, minimum width=10mm, fill=white, right=7cm of vprev] (rvprev) {$V$};
		\node[draw, minimum width=15mm, fill=white, right=0cm of rvprev] (rvdash) {$V'$};

		\node[minimum width=3mm, fill=red, left=-4mm of rvprev] {};
		\node[minimum width=1mm, fill=red, left=-9mm of rvprev] {};
		\node[minimum width=1mm, fill=red, left=-7mm of rvprev] {};
		\node[minimum width=2mm, fill=red, right=2mm of rvprev] {};
		\node[minimum width=1mm, fill=red, right=5mm of rvprev] {};
		\node[minimum width=2mm, fill=red, right=8mm of rvprev] {};
		\node[minimum width=4mm, fill=red, right=11mm of rvprev] {};

		\node[draw, dashed, minimum width=10mm, fill=white, right=7cm of xprev] (rxprev) {$X$};
		\node[draw, minimum width=15mm, fill=white, right=0cm of rxprev] (rxdash) {$X'$};

		\node[minimum width=3mm, fill=yellow, left=-4mm of rxprev] {};
		\node[minimum width=1mm, fill=yellow, left=-9mm of rxprev] {};
		\node[minimum width=1mm, fill=yellow, left=-7mm of rxprev] {};
		\node[minimum width=2mm, fill=yellow, right=2mm of rxprev] {};
		\node[minimum width=1mm, fill=yellow, right=5mm of rxprev] {};
		\node[minimum width=2mm, fill=yellow, right=8mm of rxprev] {};
		\node[minimum width=4mm, fill=yellow, right=11mm of rxprev] {};

		\draw[-{LaTeX[]}] ($(rkprev.north) + (0,2mm)$) to[bend left=60] ($(rkdash.north) + (0,2mm)$);
		
		\draw [thick, decoration={brace, mirror, amplitude=10pt, raise=0.5cm}, decorate] (xt.west |- 0,-2.75cm) -- (yt.east |- 0,-2.75cm)
		node [pos=0.5, anchor=north, yshift=-0.85cm] {(a) inference cycle};
		\draw [thick, decoration={brace, mirror, amplitude=10pt, raise=0.5cm}, decorate] ($(rvprev.west |- 0,-2.75cm) - (5mm,0)$) -- ($(rvdash.east |- 0,-2.75cm) + (5mm,0)$)
		node [pos=0.5, anchor=north, yshift=-0.85cm] {(b) memory refresh cycle};
		
	\end{tikzpicture}}
	\caption{General framework KV cache management with inference and memory refresh cycles. Our memory refresh strategy is to use counter-causal prediction as a proxy for entries to evict redundant entries from the cache. The optional buffers $X$ and $X'$ record either indexes or (input) tokens corresponding to stored keys and values.}
	\label{fig:method}	
\end{figure}
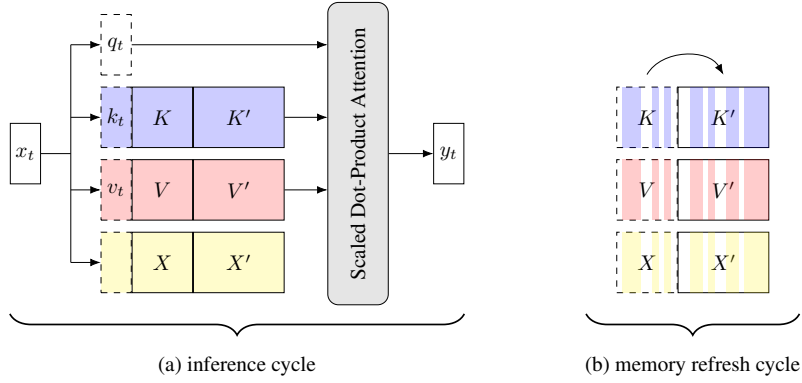

We are concerned with the task of auto-regressive token generation under limited memory constraints by large language models (LLMs) responding to instructions. Let us begin by presenting an abstract framework for key-value (KV) cache management that encompasses a variety of methods including input context compression and dynamic context eviction. Denote by $x_{1:t}$ the sequence of tokens having been processed up to time $t$ which may comprise the system prompt, the instruction, and the partial auto-regressively generated response. Assume that our model is limited to a maximum context window (i.e., token history) of length $h$ but is also equipped with an adaptive memory unit with $J$ slots.\footnote{We refer to $J$ as the cache size and $h$ as the chunk size in our description below.} In populating the memory slots, the goal is to approximate the probability distribution over the next token conditioned over the entire prior context with a distribution instead conditioned on the limited context window combined with information stored in memory. Mathematically, we want to find memory elements $z_{1:J}$ such that
\begin{align}
	P(x_{t+1} \mid x_{1:t}) &\approx P(x_{t+1} \mid x_{t-h+1:t}, z_{1:J}).
	\label{eqn:condprob}
\end{align}

Clearly searching over all possible $z_{1:J}$ is intractable and how to deal with this difficulty is where cache management strategies differ. For example, the sliding window approach places the entire memory budget into $h$ and no effective adaptive memory ($J = 0$).\footnote{Or the adaptive memory simply acts as an extension to the context history buffer as we do in our implementation.} The heavy-hitter oracle ($H_2$O)~\citep{Zhang:NIPS2023}, on the other hand, maintains a small window of recent history $h$ and chooses which subset of keys $k_i$ and values $v_i$ for $i \in \{1, \ldots, t\}$ to keep in the adaptive memory based on accumulated attention. Different from these, we use counter-causal prediction to rank candidates for eviction. Our core insight is that \emph{if the present is good at predicting the past, then we do not need to remember the past for predicting the future}.\footnote{This can be thought of as a relaxation of the classic Markov assumption, which states that the future is independent of the past given the present. Note that it is different, however, to the logical fallacy of assuming the converse, $(A \to B, B) \to A$.} This is because the model (and recent context) already captures the past information from easily predicted tokens.

Before elaborating on how this insight translates into our mechanism for populating the memory and managing the cache contents, let us first make more concrete our proposed memory augmented architecture.
For clarity of exposition we describe our method for a single head of a single attention layer. The extension to multi-head attention (MHA), group-query attention (GQA)~\citep{Ainslie:EMNLP2023}, and multiple attention layers is straightforward. Standard scaled dot-product attention~\citep{Vaswani:NIPS2017} computes,
\begin{align}
	\texttt{attn}(Q,K,V) &= \texttt{softmax}\!\left(\frac{QK\transpose}{\sqrt{d}}\right)V
\end{align}
where $Q$, $K$, and $V$ are query, key, and value matrices, respectively, with one row per input token, and where softmax is applied rowwise. Queries, keys and values are computed from each input token and from the output of successive layers of the transformer. They can be processed in batch on prompt data during the pre-fill phase or auto-regressively one at a time during response generation (also called the decode phase). In most models a causal mask is applied prior to the softmax operation to ensure that queries can only attend to past tokens when processed in batch. Moreover, keys and values, $K$ and $V$, computed from previous mini-batches (or during decode) are cached to avoid having to recompute them for each new token. For long context windows this consumes a vast amount of memory, hence the need for KV cache management.

Now let $\cM = (K', V')$ be a separate persistent memory module composed of up to $J$ editable keys and values, $K'$ and $V'$. We implement memory augmented attention as,
\begin{align}
	\texttt{mem\_aug\_attn}(Q,K,V,K',V') &= \texttt{attn}\!\left(Q, \begin{bmatrix} K' \\ K \end{bmatrix}, \begin{bmatrix} V' \\ V \end{bmatrix} \right)
\end{align}
where we have concatenated keys (resp. values) from the memory module with keys (resp. values) from the context tokens. The order in which we concatenate is arbitrary, although to maintain a causal sequence, prepending is preferred. Once again, this general setup encapsulates many methods and the concatenation $K' \oplus K$ (resp. $V' \oplus V$) can be treated as one memory block albeit populated via different mechanisms.

Note that each token in the context window can attend to both other tokens and memory slots but, unlike the tokens, there are no queries associated with memory slots, only keys and values. Indeed, during the decode phase, only the query $q_t$ associated with the latest token $x_t$ is needed. Memory slots can also be seen as providing additional contextual information without the overhead of processing additional tokens and propagating them through the transformer layers such as with register tokens~\citep{Darcet:CVPR2024}. The idea is similar to the persistent memory architecture of \citep{Sukhbaatar:ICLR2020}. However, in our approach the memory is populated from the test-time context history rather than learned during training making our method directly amenable to existing pre-trained models and zero/low-shot techniques such as in-context learning~\citep{Brown:NIPS2020}.

At each time step $t$ during the decode phase, the model constructs a query $q_t$, key $k_t$, and value $v_t$ from input $x_t$ and computes
\begin{align*}
	y_t\transpose &= \texttt{softmax}\left(\frac{q_t\transpose K\transpose}{\sqrt{d}}\right) V
\end{align*}
from which we eventually sample the next token $x_{t+1}$.%
\footnote{We have intentionally ignored subsequent projections, layer normalization, skip connections, and other architectural components for the sake of clarity. See \citet{Vaswani:NIPS2017} for full details.}
Here matrices $K$ and $V$ contain keys and values computed from time $t-h+1$ to $t$ having had key $k_t\transpose$ and value $v_t\transpose$ appended, and are understood to be prepended with up to $J$ memory slots as outlined above before performing the computation. This operation is illustrated in \figref{fig:method}(a), which we call the inference cycle.

Given this process of eventually sampling $x_{t+1}$, \eqnref{eqn:condprob} can be equivalently rewritten as
\begin{align}
	P(x_{t+1} \mid x_{1:t}) &\approx P(x_{t+1} \mid q, K, V)
\end{align}
since $q$, $K$ and $V$ are functions of $x_{1:t}$.

Due to the overhead associated with deciding which keys and values to evict from the cache and then the subsequent data structure manipulations, we operate in a chunkwise manner alternating between an inference cycle and a memory refresh cycle. The inference cycle runs the standard LLM pre-fill or decode steps where we process up to $h$ tokens, filling the $K$ and $V$ buffers as we go, and generating outputs auto-regressively during the decode phase. The memory refresh cycle is responsible for determining which out of the $h + J$ stored key-value pairs to keep and which to discard. The up to $J$ kept pairs are moved into $K'$ and $v'$, and buffers $K$ and $V$ are emptied ready for the next inference cycle as shown in \figref{fig:method}(b). Note that this procedure does not preclude recent key-value pairs being copied into the adaptive memory slots thereby maintaining short-term history in the manner of a sliding window.

\subsection{Counter-Causal Surprise}

With the high-level process outline above, we are now ready to detail our cache eviction strategy. As already introduced, if we can predict earlier tokens from later (possibly auto-regressively generated) ones, then those \emph{unsurprising} earlier tokens can be removed from the cache. To that end, we define a counter-causal surprise score for the $i$-th token as
\begin{align}
	s_i &= 1 - P(x_{i} \mid x_{i+1:t})
	\label{eq:surprise}
\end{align}

However, determining $s_i$ exactly presents us with a difficulty: computing $P(x_i \mid x_{i+1:t})$ requires reversing the direction of conditioning of standard autoregressive LLMs, which are trained to model $P(x_i \mid x_{1:i-1})$. Thus, we approximate $s_i$ by re-purposing the model's forward pass inference routine with a modified attention mask. Specifically, we run the model on the tokens in their original order, $x_1, x_2, \ldots, x_t$, with their original position IDs, but replace the standard causal (lower-triangular) mask with a counter-causal (upper-triangular) mask that allows each position $i$ to attend only to later positions $j > i$. Rather than recomputing keys and values, we reuse the key and value representations $K, V$ already stored in the KV cache from the forward pass. These cached representations carry RoPE encodings at the correct absolute positions, so computing fresh queries $Q$ with the same original position IDs gives the correct relative offsets in counter-causal attention. The output logit at position $i$ thereby approximates $P(x_i \mid x_{i+1:t})$ using the actual cached representations, directly measuring how well the remaining cache entries can compensate for the absence of $x_i$. We take the logit of $x_i$ (rather than the full softmax probability) as our surprise score, which avoids numerical issues and is a monotone function of $P(x_i \mid x_{i+1:t})$ under softmax. Since we evaluate the score on tokens already processed by the model, we store the token sequence in a small buffer $X$ alongside the KV cache; the memory overhead of integer token IDs is negligible.

During the memory refresh cycle we evaluate the counter-causal score for each token $x_i$ in buffer $X$ and keep only the keys and values associated with the $J$ highest scoring tokens.  The last token in the cache has no future context and is assigned maximum surprise (i.e., it is always retained). A one-to-one correspondence is maintained between the tokens in buffer $X$ and the keys and values in $K' \oplus K$ and $V' \oplus V$, respectively, as illustrated in \figref{fig:method}(b).

\paragraph{Fast single-layer approximation.}
The full counter-causal pass described above has cost $O(L n^2)$ where $L$ is the number of transformer layers and $n$ is the current cache size. We propose a much cheaper approximation that reduces this to $O(n^2)$ by restricting the counter-causal pass to the last transformer layer only. During the standard causal forward pass we record $H^{(L-1)}$, the hidden activations entering the last transformer layer. This consumes an additional $nd$ values that need to be stored, or about 12\% of KV cache memory budget for Qwen2.5-7B and dropping to 5\% for Qwen2.5-14B. During a memory refresh we compute queries $Q$ from the stored activations $H^{(L-1)}$ and apply RoPE with the original position IDs. Attention is computed using a counter-causal (upper-triangular) mask so that each position $i$ attends only to positions $j > i$ later in time. Finally, we apply the layer's standard output processing to obtain logits for each token in the cache. This fast approximation costs a single attention layer rather than $L$ full transformer layers. On Qwen2.5-7B ($L=28$ layers), the measured refresh latency drops from 54ms to 7.9ms at $n=512$ and from 496ms to 52.6ms at $n=4096$, a speedup of 7--9 times per refresh cycle (Table~\ref{tab:efficiency}).

\subsection{Algorithm Summary}

The following pseudo-code summarizes our approach. Note that in the pre-fill phase (Lines 3--8) the inference cycle does not need to predict the next token and the input can be processed in batches of $h$ tokens. In contrast, the decode phase (Lines 10--15) auto-regressively generates the next token $x_{t+1}$ one at a time until a maximum number of tokens is produced or some other termination criteria is satisfied (e.g., generation of an end-of-sequence token).

\begin{framedalgo}
\begin{algorithmic}[1]
	\Function{CacheManagedInference}{$x_{1:T}$, $J$, $h$}
	\Comment Prompt, Cache Size, Chunk Size
		\State initialize memory $K$, $V$, $\cM = (K', V')$
		\Comment Clear Buffers
		\For{$t = 1, \ldots, T-1$}
			\Comment Pre-Fill Phase
			\State \_ = \Call{DoInference}{$x_{t}$}
			\Comment Inference Cycle
			\If{$t \,\%\, h = 0$}
				\State \Call{DoMemRefresh}{ }
				\Comment Memory Refresh Cycle
			\EndIf
		\EndFor
		\State
		\For{$t = T, \ldots, \text{max\_tokens}$}
			\Comment Decode Phase
			\State $x_{t+1}$ = \Call{DoInference}{$x_{t}$}
			\Comment Inference Cycle
			\If{$t \,\%\, h = 0$}
				\State \Call{DoMemRefresh}{ }
				\Comment Memory Refresh Cycle
			\EndIf
		\EndFor
	\EndFunction
\end{algorithmic}
\end{framedalgo}

For completeness, pseudo-code for the inference and memory refresh steps is also given. Similar to the pre-fill forward pass, the memory refresh cycle can process tokens in batch even during decode since no new tokens are sampled (Lines 2--4).

\begin{framedalgo}
\begin{algorithmic}[1]
	\Function{DoInference}{$x_t$}
		\State compute queries, keys and values, $q_t, k_t, v_t = f(x_t)$
		\State $K \gets K \oplus k_t$
		\Comment Update KV Cache
		\State $V \gets V \oplus v_t$
		\State $X \gets X \oplus x_{t}$
		\Comment Remember Token
		\State $\tilde{x}_t \sim P(\tilde{x} \mid q_t, K' \oplus K, V' \oplus V)$
		\Comment Sample Next Token
		\State \Return $\tilde{x}_t$
	\EndFunction	
\end{algorithmic}
\end{framedalgo}

\begin{framedalgo}
	\begin{algorithmic}[1]
		\Function{DoMemRefresh}{ }
		\For{$t = 1, \ldots J + h$}
			\State $s_t = 1 - P(X_t \mid X_{t+1:J+h}$)
			\Comment Counter-Causal Surprise
		\EndFor
		\State $\sigma$ = \textbf{argsort}($s_1, \ldots, s_{J+h}$)
		\Comment Sort Largest to Smallest
		\State $K' \gets (K' \oplus K)_{\sigma[1:J]}$
		\Comment Update memory
		\State $V' \gets (V' \oplus V)_{\sigma[1:J]}$
		\State $X \gets X_{\sigma[1:J]}$
		\State clear $K$ and $V$
		\EndFunction	
	\end{algorithmic}
\end{framedalgo}
	
The pseudo-code above shows \emph{chunked} refresh mode where \textsc{DoMemRefresh} is called every $h$ tokens in both the pre-fill and decode phases. We could also run in \emph{prefill-end} mode where the refresh hook fires exactly once at the end of the pre-fill phase, before the decode step begins. This is the behavior of offline cache compression techniques, but is less relevant for cache management of during long-context generation tasks. For the fast approximation, the counter-causal probability in \textsc{DoMemRefresh} is computed via the single-layer approach described above rather than a full $L$-layer counter-causal pass.

In practice, for tasks with long input contexts, such as LongHealth~\citep{adams2024longhealth}, Qasper~\citep{Dasigi:NAACL2021}, and LoCoMo~\citep{maharana2024lococmo}, we also keep the system prompt defining the task (but not the specific full context for the instance being tested) in memory. This consumes a small number of our $J$ memory slots, which are frozen so that they cannot be evicted from the cache. These act as sink tokens, which have been found important in other KV cache eviction works~\cite{Xiao:ICLR2024}. We do this for all strategies being tested. The system prompts for tasks explored in our experiments are included in \appref{sec:prompts}.

\section{Experiments}
\label{sec:expr}

We conduct experiments on four benchmark tasks using four open-weight large language models from two families: Qwen2.5 (3B, 7B, and 14B Instruct variants) and LLaMA 3.1 (8B Instruct). Our interest is in comparing the quality of model output under different cache management strategies relative to a model with no limit on the size of the cache rather than absolute performance on the task.

The cache management strategies that we compare are:
\begin{description}[labelindent=2em,leftmargin=10em,style=nextline]
	\item[Full:] Runs the model with no limit on the cache size to provide a baseline for comparison. However, a cap is placed on the maximum number of tokens that can be generated, which is task specific but the same for all strategies. 
	\item[Sliding window:] Sometimes called local, this is a standard baseline where the KV cache is implemented as a first-in first-out queue. Once the cache is full the oldest entries are evicted to make room for new entries.
	\item[Importance:] A simplified version of the heavy-hitter oracle~\citep{Zhang:NIPS2023} and similar in spirit to TOVA~\citep{Oren:arXiv2024} where tokens with lowest mean attention score in the model's last layer are evicted. Different to TOVA, we use output layer keys as a proxy for queries.
	\item[Heavy-hitter:] The heavy-hitter oracle ($H_2$O)~\citep{Zhang:NIPS2023} computed as running total of accumulated attention scores summed over all layers for each token. Uses the most recent keys as a proxy for queries. Non-recent tokens with the lowest accumulated scores are evicted.
	\item[Counter-causal:] Our method that performs counter-causal prediction. Tokens with the lowest surprise score (i.e., most predictable) are evicted from the cache. We evaluate two variants: \emph{counter} runs a full $L$-layer pass, reusing the existing cached keys and values with a counter-causal mask, and \emph{counter (fast)}, which restricts the counter-causal pass to the last transformer layer only.
\end{description}

We include sliding window and $H_2$O as baselines, representing single-pass recency- and attention-based eviction, respectively. PyramidKV~\citep{Cai:arXiv2024} and Ada-KV~\citep{Feng:arXiv2024} modulate the cache budget per layer but use the same forward-attention signal as $H_2$O and are thus vulnerable to the same self-reinforcing bias; we discuss them in related work rather than include them as baselines.

Our benchmark tasks were chosen to test both long input contexts and long-form responses. For example, MATH500~\citep{Hendrycks:NIPS2021} requires that a model solve a given mathematics problem using chain-of-thought reasoning, which results in long sequences of tokens generated in the output stage before the final answer is given. In contrast, LongHealth~\citep{adams2024longhealth} tests how well LLMs can read, reason over, and extract information from long clinical records, but only requires short answers to a multiple-choice question. Qasper~\citep{Dasigi:NAACL2021} similarly has long input context but in this case for the scientific domain. Answers are typically short. Finally, LoCoMo~\citep{maharana2024lococmo} is a dataset of very long conversations between two speakers, each comprising 300 turns over multiple sessions. The LLM is required to answer a question using exact phrases from the conversation, but the responses are once again typically short.

All experiments use greedy decoding (i.e., one rollout per problem). Some evaluation protocols in the literature call for the use of a second LLM to evaluate responses. However, since our interest is performance relative to the model with unlimited memory ({\sc Full}) and we do not want to confound results with the reasoning abilities of a second model, we elect to use the standard accuracy metric for MATH500 and LongHealth requiring an exact match, and the mean F1-score for Qasper and LoCoMo where precision and recall are computed by comparing the set of generated tokens to a ground truth answer.

Accuracy experiments (Table~\ref{tab:math500res} and Figures~\ref{fig:results},~\ref{fig:math500len},~\ref{fig:math500cache}) were conducted on a single NVIDIA H100 NVL GPU with 94GB of onboard memory using Python 3.12 and PyTorch \texttt{2.10.0+cu128}.
Efficiency benchmarks (Table~\ref{tab:efficiency}) were conducted on an NVIDIA RTX~4090 GPU using Python 3.11 and PyTorch \texttt{2.10.0+cu128}.
Data and models were obtained from Hugging Face or their original source.

\subsection{Efficiency}
\label{sec:efficiency}

\begin{table}[h]
\centering
\small
\setlength{\tabcolsep}{5pt}
\begin{tabular}{l|cc|cc}
\hline
 & \multicolumn{2}{c|}{\sc Refresh latency (ms)} & \sc E2E time & \sc Peak GPU \\
\sc Method & $n{=}512$ & $n{=}4096$ & (s / sample) & memory (GB) \\
\hline
Full (no eviction) & --- & --- & 10.6 & 15.4 \\
Sliding window     & $<\!1$ & $<\!1$ & 12.2 & 15.4 \\
Heavy-hitter (H2O) & $<\!1$ & $<\!1$ & 10.5 & 15.4 \\
Counter-causal (ours) & 54 & 496 & 10.2 & 16.0 \\
Counter-causal (fast, ours) & 7.9 & 52.6 & 10.2 & 15.6 \\
\hline
\end{tabular}
\smallskip
\caption{Per-refresh latency and end-to-end time on Qwen2.5-7B (RTX 4090, fp16). Refresh latency is measured as the mean over five runs with a synthetic KV cache of the stated size $n$. End-to-end time is per-sample on 50 MATH500 problems (cache size $J=512$, chunk size $h=256$). Peak memory is the maximum GPU allocation across all 50 samples.}
\label{tab:efficiency}
\end{table}

Table~\ref{tab:efficiency} reports per-refresh latency for both of our counter-causal variants alongside attention-based eviction baselines. Our first approach runs a full $L$-layer counter-causal pass (54ms at $n=512$; 496ms at $n=4096$); the fast variant restricts the pass to the last layer (7.9ms; 52.6ms), resulting in a 7--9 times reduction for each refresh cycle. The end-to-end (E2E) times show that limiting the KV cache size to $J=512$ actually reduces per-token attention cost during the long decode phase, making both counter-causal variants (10.2 seconds per sample) comparable to the full-cache baseline (10.6s) despite the refresh overhead. The higher E2E for sliding window (12.2s) reflects its lower accuracy on these 50 problems, since failed attempts generate longer, incorrect reasoning chains before terminating. The real benefit of the fast approximation appears in long-context settings (e.g., LoCoMo and LongHealth) where the sequence length can reach 4096--24576 tokens and the refresh cycle run many times per prompt. At $n=4096$ the fast approach adds only 52.6ms per refresh call compared to 496ms for the full pass and less than 1ms for attention-based methods. Counter (fast) stores a small $H^{(L-1)}$ buffer (approximately 12\% of the KV cache budget for Qwen2.5-7B), accounting for the modest increase in peak GPU memory (15.6GB versus 15.4GB for baselines). Our primary counter-causal method uses no such buffer but incurs higher peak memory (16.0GB) from intermediate activations during the full $L$-layer refresh pass.

\subsection{Math500}

\begin{table}
\centering
\small
\setlength{\tabcolsep}{5pt}
\begin{tabular}{l|c|ccccc}
\hline
\sc Instruct & & & \sc Import- & & \sc Counter- & \sc Counter \\
\sc Model & \sc Full & \sc Sliding & \sc ance & \sc H2O & \sc Causal & \sc (fast) \\
\hline
Qwen2.5-3B   & 0.652 & 0.512 & 0.522 & 0.526 & {\bf 0.602} & 0.592 \\
Qwen2.5-7B   & 0.766 & 0.692 & 0.692 & {\bf 0.762} & 0.744 & 0.736 \\
Qwen2.5-14B  & 0.814 & 0.654 & 0.648 & 0.646 & {\bf 0.758} & 0.740 \\
Llama-3.1-8B & 0.488 & 0.458 & 0.458 & 0.464 & {\bf 0.482} & 0.480 \\
\hline
\end{tabular}
\smallskip
\caption{Performance on the MATH500~\cite{Hendrycks:NIPS2021} benchmark. Cache size $J=512$, chunk size $h=256$, maximum output 2048 tokens. Counter-causal is our primary method; counter (fast) uses the single-layer approximation. Bold indicates the best eviction method per row.}
\label{tab:math500res}
\end{table}

\begin{figure}[h]
\centering
\includegraphics[width=0.4\textwidth]{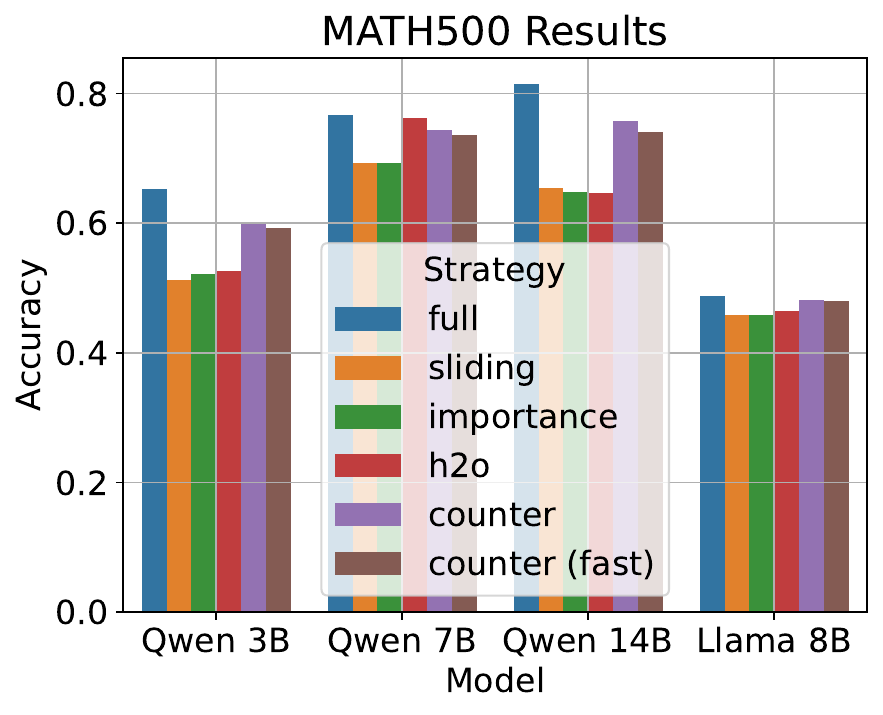}
\caption{MATH500 accuracy by model and eviction strategy (same data as Table~\ref{tab:math500res}).}
\label{fig:math500bar}
\end{figure}

Results for MATH500~\citep{Hendrycks:NIPS2021} are shown in \tabref{tab:math500res}. Our counter-causal method is the best performing eviction method on three of the four models, achieving 60.2\%, 74.4\%, and 75.8\% on Qwen2.5-3B, 7B, and 14B respectively. On Qwen2.5-7B, $H_2$O leads among the eviction methods (76.2\% versus our 74.4\%). The fast single-layer approximation (Section~\ref{sec:main}) achieves 73.6\% on Qwen2.5-7B, slightly below the full counter-causal method at 74.4\% but with an eight times lower refresh cost (see Table~\ref{tab:efficiency}). Interestingly, performance for the Qwen2.5 model family is vastly superior to that of Llama3.1, which suggests that Qwen has specifically been tuned to answer mathematics questions (or has seen some of the MATH500 data during training). However, this observation is immaterial for the comparison of the different eviction strategies, where we only care about relative performance to the {\sc Full} baseline.

On Llama-3.1-8B, our method achieves 48.2\%, which is only slightly below the full-cache baseline of 48.8\%, and remains the best of all eviction methods tested.

\begin{figure}
	\centering
	\begin{tabular}{ccc}
		\includegraphics[width=0.4\textwidth]{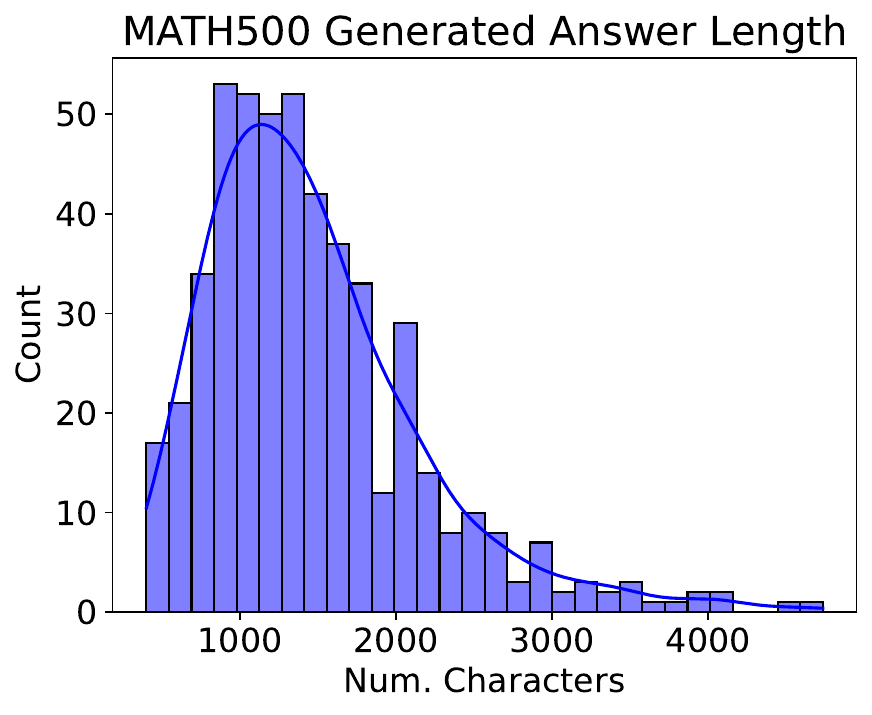} 
		& \hspace{0.05\textwidth} &
		\includegraphics[width=0.4\textwidth]{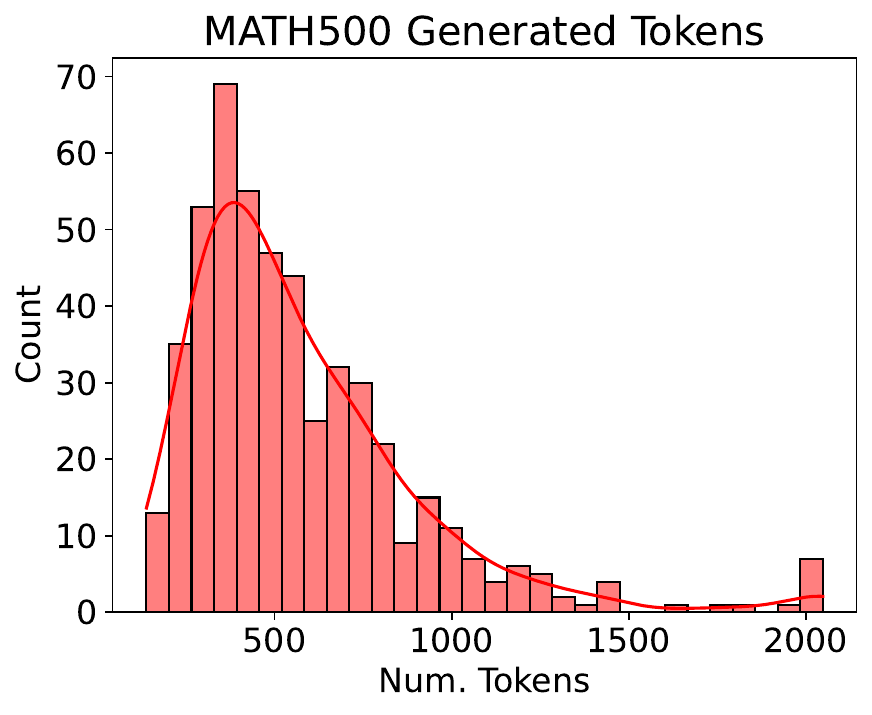}
	\end{tabular}
	\caption{Histogram of generated output sequence length by number of characters (left) and number of tokens (right) for the problems in the MATH500~\citep{Hendrycks:NIPS2021} dataset on Qwen-2.5-7B-Instruct using full input context.}
	\label{fig:math500len}	
\end{figure}

\begin{figure}
	\centering
	\setlength{\tabcolsep}{3pt}
	\begin{tabular}{ccc}
		\includegraphics[width=0.31\textwidth]{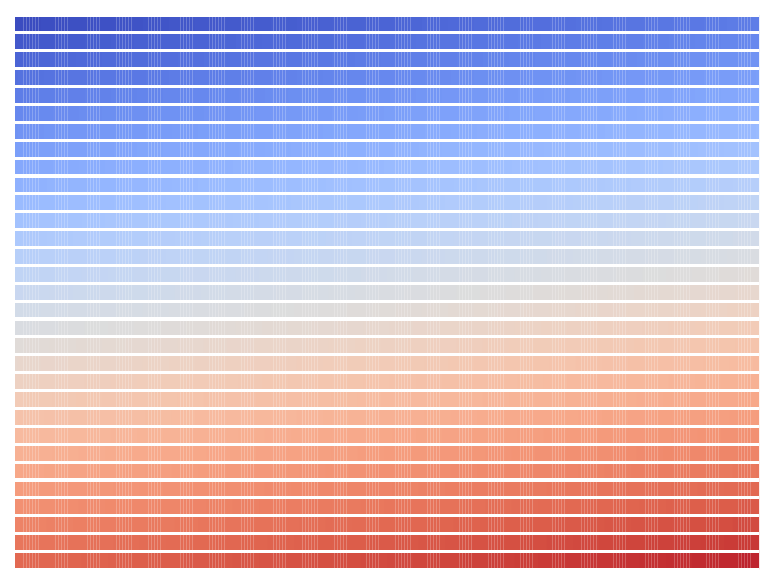} &
		\includegraphics[width=0.31\textwidth]{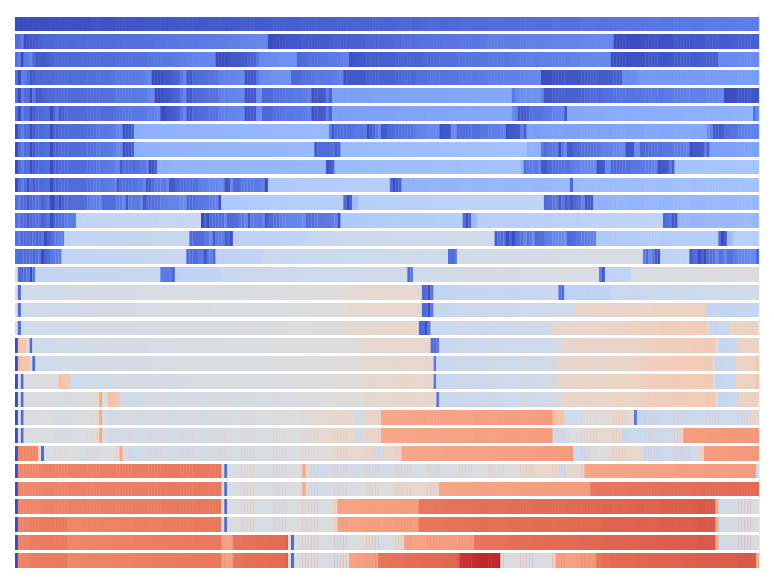} &
		\includegraphics[width=0.31\textwidth]{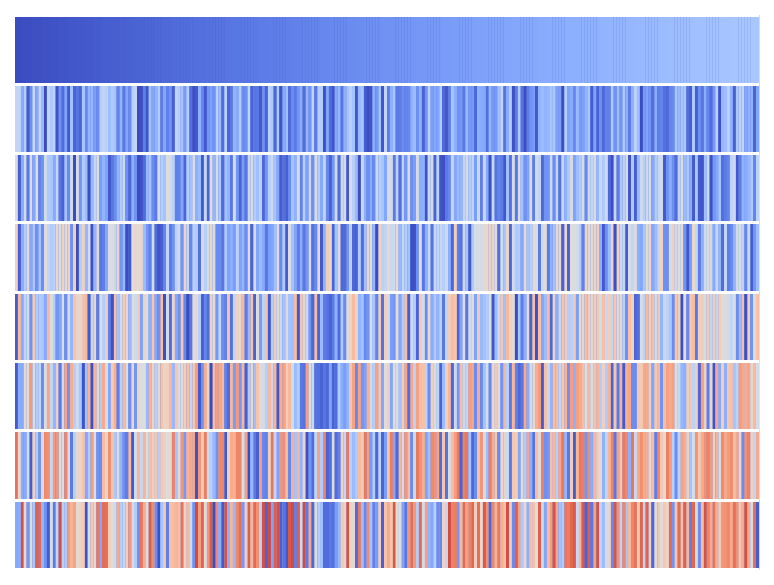}
		\\
		{\small (a) sliding window} &
		{\small (b) importance sampling} &
		{\small (c) counter-causal}
	\end{tabular}
	\caption{Visualization of the memory augmented key-value cache for different strategies on MATH500 Problem \#2. Rows show evolution of cache after each refresh cycle. Columns indicate cache entry colored by generation timestep (token position) in the output sequence. Not only does our counter-causal cache management strategy improve accuracy, but also, in this example, produces a shorter answer (and hence uses less compute) as can be seen by the fewer number of rows.}
	\label{fig:math500cache}
\end{figure}

To gain a better understanding of the results we plot a histogram of the output sequence length in \figref{fig:math500len}. The plot shows that for the majority of question, the generated answer comes after 256 tokens, and many problems require one or more memory refresh cycles. We also visualize the contents of the KV cache for one of the problems from the dataset in \figref{fig:math500cache}. Each row depicts the cache after a refresh cycle. Here we set the chunk size to 64 for illustrative purposes, i.e., rows are updated every 64 tokens.

\subsection{AIME (Thinking-Mode Reasoning)}
\label{sec:aime}

To evaluate cache eviction in an even more decode-heavy settings, we benchmark on 90 problems from the AIME 2022--2024 competition mathematics dataset~\citep{Dekoninck:arXiv2026} using Qwen3-8B in thinking mode.\footnote{The dataset is available from HuggingFace as \texttt{AI-MO/aimo-validation-aime}.}. Thinking-mode models generate an extended chain-of-thought reasoning trace (enclosed in \texttt{<think>...</think>} tags) before producing a final boxed answer. This yields output sequences of up to 16384 tokens, far longer than the standard MATH500 setting, thereby creating sustained pressure on the KV cache throughout the decode phase.

We use a chunk size of $h=2048$ and cache budget of $J=4096$ (25\% retention at 16k). The proportion of problems with no extractable answer ({\tt pred=None}) due to exceeding the token limit before the \texttt{</think>} tag closes provides an additional diagnostic. Specifically, it indicates whether the model's reasoning chain is being disrupted by evictions.

\begin{table}
\centering
\small
\begin{tabular}{l|c|ccccc}
	\hline
	& & & \sc Import- & & \sc Counter- & \sc Counter \\
	& \sc Full & \sc Sliding & \sc ance & \sc H2O & \sc Causal & \sc (fast) \\
	\hline
	{\sc Accuracy} & 0.500 & 0.267 & 0.144 & 0.333 & {\bf 0.367} & 0.289 \\
	{\tt pred=None} & 46\% & 70\% & 83\% & 58\% & 57\% & 66\% \\
	\hline
\end{tabular}
\smallskip
\caption{AIME accuracy on Qwen3-8B (thinking mode). Cache size $J=4096$ and chunk size $h=2048$. The full baseline correctly answers 50.0\% (45/90) of problems. Row ``{\tt pred=None}'' indicates failure to complete the reasoning chain within the 16k token limit.}
\label{tab:aime}
\end{table}

Results are shown in Table~\ref{tab:aime}. The full-cache baseline achieves 50.0\%, with 46\% of problems producing no answer (an irreducible floor representing problems that require more than 16384 output tokens even without compression). At a cache size limit of 4096 (25\% retention and up to 64 memory refresh cycles over a 16k decode window), all methods degrade substantially. Frequent evictions disrupt the reasoning chain, preventing the model from closing the \texttt{</think>} tag and producing a final answer. Our counter-causal approach at 36.7\% accuracy is the best performing method at this budget, confirming its ability to preserve reasoning coherence more than other methods.

\subsection{LongHealth, Qasper, and LoCoMo}

\begin{figure}
	\centering
	\small
	\setlength{\tabcolsep}{0pt}
	\begin{tabular}{ccc}
		\includegraphics[width=0.33\textwidth]{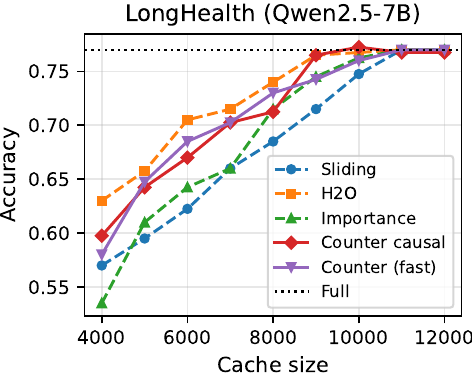} &
		\includegraphics[width=0.33\textwidth]{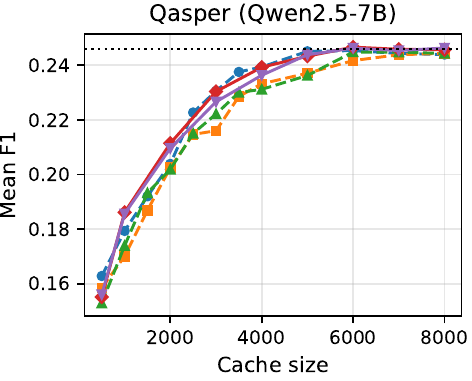} &
		\includegraphics[width=0.33\textwidth]{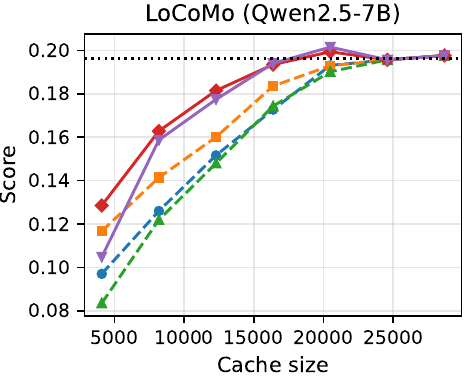}
		\\
		\includegraphics[width=0.33\textwidth]{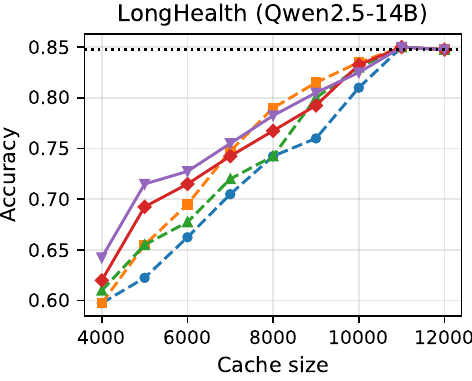} &
		\includegraphics[width=0.33\textwidth]{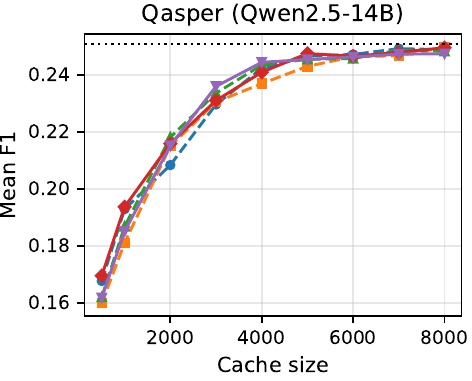} &
		\includegraphics[width=0.33\textwidth]{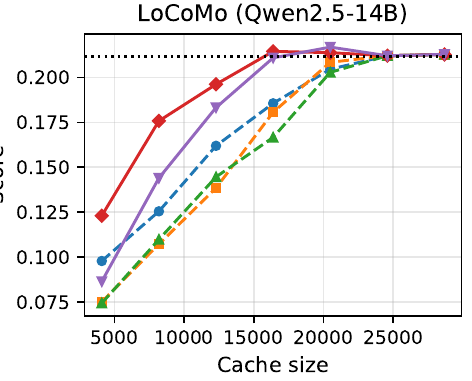}
		\\
		\includegraphics[width=0.33\textwidth]{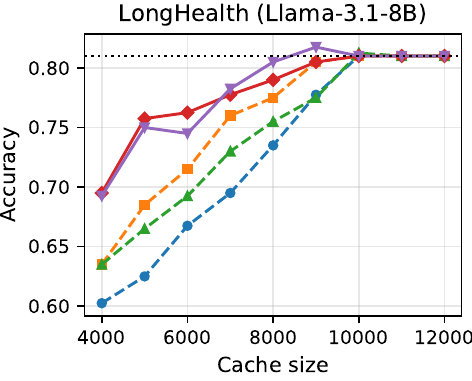} &
		\includegraphics[width=0.33\textwidth]{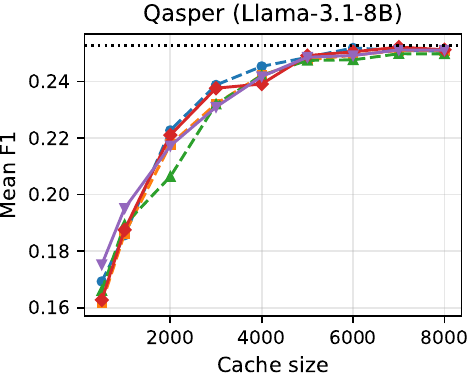} &
		\includegraphics[width=0.33\textwidth]{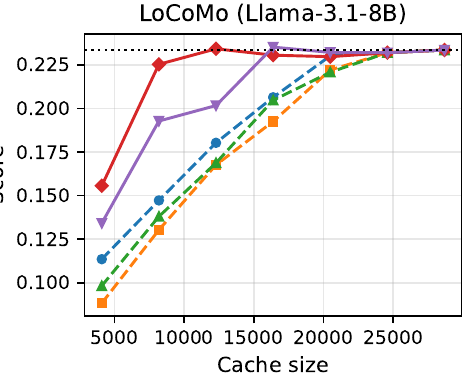}
		\\
		(a) LongHealth &
		(b) Qasper &
		(c) LoCoMo
		\\
		\citep{adams2024longhealth} &
		\citep{Dasigi:NAACL2021} &
		\citep{maharana2024lococmo}
	\end{tabular}
	\caption{Experimental results on three different long context tasks as a function of cache size comparing different cache eviction strategies and pre-trained large language models.}
	\label{fig:results}
\end{figure}

Results on LongHealth~\citep{adams2024longhealth}, Qasper~\citep{Dasigi:NAACL2021} and LoCoMo~\citep{maharana2024lococmo} are shown in \figref{fig:results}. The LoCoMo benchmark consists of 1,986 questions divided into five categories. We evaluate only on the multi-hop category consisting of 282 questions, which requires synthesizing information across multiple sessions. For the LongHealth and Qasper tasks we include all questions from the evaluation dataset, 400 and 1005, respectively.  

In these experiments we sweep over cache sizes, setting the chunk size (memory refresh cycle rate) to 25\% of the cache size. The horizontal dashed line represents task performance with no memory limit, which can also be seen as where the different strategies converge at the extreme right of each plot.

As expected all strategies show degraded performance as the cache size is reduced. Our counter-causal strategy outperforms the other methods for most tasks. For LongHealth on Qwen2.5-7B the performance of the counter-causal strategy and heavy-hitter ($H_2$O) strategy are equal across the different cache sizes, but counter-causal clearly outperforms $H_2$O and other strategies on Llama3.1-8B. Interestingly, the fast approximation sometimes outperforms our full counter-causal method. On Qasper all strategies perform about the same across all models.

Interestingly, $H_2$O struggles on LoCoMo for small cache sizes on both Qwen2.5-14B and Llama3.1-8B models. We analyzed these small cache responses and found two characteristic failure modes. First, the $H_2$O strategy output literally repeats the input question as the answer, suggesting that the model has lost all relevant conversational context. Second, the method injects irrelevant image captions from the conversation, indicating that high-attention image-related tokens are retained at the expense of factual content.
We attribute this to the self-reinforcing nature of attention-based eviction and hypothesize that unique facts mentioned once across a large tokens sequence receive little cumulative attention during the pre-fill phase and are evicted early. Once evicted, the model cannot answer questions about them and fixates on the most-attended recent context (the question itself) or on image captions that attracted attention from multiple nearby tokens. Counter-causal eviction avoids this failure mode precisely because it measures how \emph{surprising} a token is given its future context: a unique fact that cannot be inferred from surrounding tokens has high surprise and is protected from eviction even if it was rarely attended.

\section{Conclusion and Discussion}
\label{sec:conclusion}

In this paper we propose a novel eviction strategy for key-value (KV) cache memory management in large language models. Our strategy is based on counter-causal surprise, which measures the ability for a model to predict a previous token given the tokens that follow it in the cache. Tokens that cannot be predicted with high probability have a high surprise score and represent unique information not available from other tokens or combinations thereof. Conversely, tokens that can be easily predicted have low surprise and offer little information not already contained in later tokens. Hence, key-value pairs associated with high counter-causal surprise tokens should be kept whereas those associated with low counter-causal surprise may be dropped.

We implement our strategy in an inference algorithm that cycles between processing tokens and refreshing the contents of the KV cache to satisfy memory limits. Our scoring pass runs in the original token order, reusing the existing cached key and value representations with a counter-causal (upper-triangular) attention mask. This keeps the computation in-distribution and tied to the actual cache contents. To further reduce cost, we additionally propose a fast single-layer approximation that restricts the counter-causal pass to the last transformer layer, reducing the per-refresh cost and achieving a 7--9 times speedup with only small accuracy loss. Our strategy performs well against alternative cache eviction methods in experiments on a variety of long context tasks requiring eviction during both the pre-fill phase and decode phase.

{\bf Limitations.}
Our primary counter-causal method requires an additional $O(L n^2)$ computation per refresh cycle, roughly doubling inference cost during pre-fill when refreshes are frequent. However, during the autoregressive decode phase, the cost is amortized over a batch of tokens and negligible compared to cost of the one-token-at-a-time inference cycle. The fast single-layer approximation reduces this to $O(n^2)$ at a small accuracy cost and the additional overhead of storing penultimate layer activations. Counter-causal surprise is an approximation of the true quantity of interest since neither the cached forward pass nor the single-layer proxy exactly models $P(x_i \mid x_{i+1:t})$. Despite this approximation, counter-causal surprise provides a principled measure of token information content that does not suffer from the self-reinforcing bias of attention-based scores.

Token-level scoring is the natural choice for a token-level eviction strategy. However, some individual tokens may be easily predicted from nearby context causing counter-causal surprise to underestimate their importance on exact retrieval tasks and evict them from the cache. Addressing this with coarser grained eviction strategies (e.g., at the phrase or sentence level) or hybrid approaches that combine counter-causal scoring with recency or retrieval-aware components is an interesting direction for future work.


\bibliographystyle{plainnat}
\bibliography{long,papers}


\newpage
\appendix

\section{Benchmark Prompts}
\label{sec:prompts}

\lstset{language=bash,numbers=none,numbersep=5pt,tabsize=4,basicstyle=\footnotesize\ttfamily}

\subsection{MATH500}

\subsubsection{Full Prompt}
\begin{lstlisting}
Solve the following math problem. Show your reasoning step by step,
then put your final answer in \\boxed{{}}.
    
Problem: {problem}
	
Solution:
\end{lstlisting}

\subsection{LongHealth}
\subsubsection{System Prompt}
\begin{lstlisting}
Read the following patient records and answer the multiple-choice question by
responding with only the letter of the correct answer (A, B, C, D, or E).
\end{lstlisting}

\subsubsection{Full Prompt}
\begin{lstlisting}
Read the following patient records and answer the multiple-choice question by
responding with only the letter of the correct answer (A, B, C, D, or E).

Patient Records:
{context}

Question: {question}

A) {answer_a}
B) {answer_b}
C) {answer_c}
D) {answer_d}
E) {answer_e}

Answer:
\end{lstlisting}

\subsection{Qasper}
\subsubsection{System Prompt}
\begin{lstlisting}
You are an expert researcher. Read the following scientific paper carefully.
Answer the user's question based only on the provided text. Responses can be
extractive, abstractive or yes/no.
	
1. Identify the relevant paragraphs, tables, or figures (Evidence).
2. Formulate an answer based on the evidence.
3. If the answer cannot be found in the text, respond with "Unanswerable".
\end{lstlisting}

\subsubsection{Full Prompt}
\begin{lstlisting}
You are an expert researcher. Read the following scientific paper carefully.
Answer the user's question based only on the provided text. Responses can be
extractive, abstractive or yes/no.

1. Identify the relevant paragraphs, tables, or figures (Evidence).
2. Formulate an answer based on the evidence.
3. If the answer cannot be found in the text, respond with "Unanswerable".

Title: {title}
Abstract:
{abstract}
	
{sections}

Question: {question}

Answer:
\end{lstlisting}

\subsection{LoCoMo}
\subsubsection{System Prompt}
\begin{lstlisting}
You are an AI assistant tasked with analyzing a conversation between
{speaker1} and {speaker2}.
Based on the provided conversation sessions, answer the question accurately.
Focus on recalling past facts, user preferences, and temporal relationships.
Answer the question using exact words from the conversation when possible.
\end{lstlisting}

\subsubsection{Full Prompt}
\begin{lstlisting}
You are an AI assistant tasked with analyzing a conversation between
{speaker1} and {speaker2}.
Based on the provided conversation sessions, answer the question accurately.
Focus on recalling past facts, user preferences, and temporal relationships.
Answer the question using exact words from the conversation when possible.

Input: {conversation}

Question: {question}

Answer:	
\end{lstlisting}

\end{document}